\documentclass{article}

     \usepackage[preprint]{neurips_2021}

\usepackage[utf8]{inputenc} 
\usepackage[T1]{fontenc}    
\usepackage{hyperref}       
\usepackage{url}            
\usepackage{booktabs}       
\usepackage{amsfonts}       
\usepackage{nicefrac}       
\usepackage{microtype}      
\usepackage{xcolor}         
\usepackage{amsmath}
\usepackage{graphicx}
\usepackage[ruled]{algorithm2e}
\usepackage{natbib}
\usepackage{bookmark}
\usepackage{xcolor}
\usepackage{amsmath}
\usepackage{svg}
\usepackage{wrapfig}
\usepackage{subcaption}
\usepackage{listings}
\usepackage{tikz}
\usepackage{tabto}

\newtheorem{definition}{Definition}[section]

\definecolor{bluecite}{HTML}{0875b7}

\hypersetup{
  colorlinks=true,
  citecolor=bluecite,
  linkcolor=magenta
}

\definecolor{codegreen}{rgb}{0,0.6,0}
\definecolor{codegray}{rgb}{0.5,0.5,0.5}
\definecolor{codepurple}{rgb}{0.58,0,0.82}
\definecolor{backcolour}{rgb}{0.95,0.95,0.92}

\SetAlgorithmName{Block}{List of protocols}

\lstdefinestyle{mystyle}{
    backgroundcolor=\color{backcolour},   
    commentstyle=\color{codegreen},
    keywordstyle=\color{magenta},
    numberstyle=\tiny\color{codegray},
    stringstyle=\color{codepurple},
    basicstyle=\ttfamily,
    breakatwhitespace=false,         
    breaklines=true,                 
    captionpos=b,                    
    keepspaces=false,                 
    numbers=left,                    
    numbersep=5pt,                  
    showspaces=false,                
    showstringspaces=false,
    showtabs=false,                  
    tabsize=2
}

\usepackage{soul} 

\title{Causal Multi-Agent Reinforcement Learning: \\ Review and Open Problems}

\author{%
    St John Grimbly\thanks{Work done during an internship at InstaDeep Ltd. Correspondence: me@stjohngrimbly.com} \\
    University of Cape Town \\
    \And
    Jonathan Shock \\
    University of Cape Town \\
    \And
    Arnu Pretorius\\
    InstaDeep \\
}

\begin{document}
\maketitle

\begin{abstract}
This paper serves to introduce the reader to the field of multi-agent reinforcement learning (MARL) and its intersection with methods from the study of causality. We highlight key challenges in MARL and discuss these in the context of how causal methods may assist in tackling them. We promote moving toward a `causality first' perspective on MARL. Specifically, we argue that causality can offer improved safety, interpretability, and robustness, while also providing strong theoretical guarantees for emergent behaviour. We discuss potential solutions for common challenges, and use this context to motivate future research directions.
\end{abstract} 
  
\section{Introduction}
In recent years there has been increasing interest in the possibilities offered to machine learning through a better understanding of causality. \cite{ref:SevenTools_Pearl} argues that explicit causal modelling in AI is crucial for achieving general intelligence. The \emph{ladder of causation} is a meta-model which expresses how different types of interaction with a data-generating system (agent-environment interface) limits the types of reasoning an agent can perform. In this sense, reinforcement learning (RL) is seen as a more powerful approach than conventional machine learning methods where purely observational data are used for regression or classification. Despite being able to perform interventions (i.e. take actions) in the environment, without a formal causal model, RL agents lack the ability to explicitly reason counterfactually. Related arguments have prompted interest in the possibility of bridging RL with methods from causality by reformulating RL models/paradigms. This has been attempted with specific tasks in mind, such as off-policy learning \citep[][]{ref:CF-GPS}, data-fusion \citep[][]{ref:BareinboimPearl-DataFusion,ref:CRL-with-DataFusion,ref:FPB-2017}, and counterfactual reasoning \citep[][]{ref:FB-2019}. This paper considers extending causal RL methods to the multi-agent case, where additional complexities arise due to interacting agents. Further, it posits that causal tools offer appropriate properties for solving some of these challenges.

\textbf{Reinforcement Learning}. The RL problem concerns that of how to map states to actions so as to maximise a numeric reward signal \citep[][]{ref:SuttonBarto, ref:Bertsekas, ref:DeepRLLectures}. An RL agent attempts to learn to select an optimal sequence of actions by trial-and-error. This reward signal may be partially observed, noisy, or confounded by any number of factors. The difficulty of the learning problem can be exacerbated by stochasticity or non-stationarities in the environment. A useful model for sequential decision making scenarios is defined by the Markov Decision Process (MDP) \citep{ref:MDP_Definition_Bellman}.

\begin{definition}[Markov Decision Process (MDP)] 
\label{def:MDP}
    A Markov decision process (MDP) is a stochastic process in which rewards are obtained when transitioning to a new state, dependent on the previous state and the action selected. Formally, it is a 5-tuple $(S,A,T,R,\gamma)$ of states $S$, actions $A$, transition probability function $T = P(s^\prime \mid s,a)$, rewards $R$, and a discount factor $\gamma$.
\end{definition}


The Markov property implies that the history of how the agent arrived at the current state is irrelevant. This means an optimal policy should prescribe the same action in a particular state regardless of the state-action trajectory. In some problems, agents have limited access to full state information. This is commonly modelled as a Partially-Observable MDP (POMDP) \citep{ref:POMDP_Definition}. Naturally, the Markov assumption does not hold in general. Consider the \emph{dynamic treatment regime} (DTR) where an individual receives a treatment plan (policy) based on their medical history and unique characteristics \citep{ref:Optimal_DynamicTreatmentRegimes, ref:DTRs_QLearning}. In a DTR, a doctor should make use of all available information to optimise long term patient outcomes. DTRs are made more difficult by the presence of latent (unobserved) factors that influence variables a healthcare worker has access to. In the case of patient outcomes, there is no way to know the exact effect an intervention will have on an individual over a long time period due to the complexity of the system. For example, remission due to a multiple-course chemotherapy treatment plan could be dependent on initial treatment \citep{ref:DTR_Chemo}. This relationship could be determined by unknown causal factors, in which case no single state would satisfy the Markovian assumption. In general, there may be information or prior knowledge about variables in a system that forms a model over the assumptions. We now discuss ways to formalise causal reasoning within a system of complex interactions and assumptions.

\begin{figure}[!t]
    \centering
    \includegraphics[width=\linewidth]{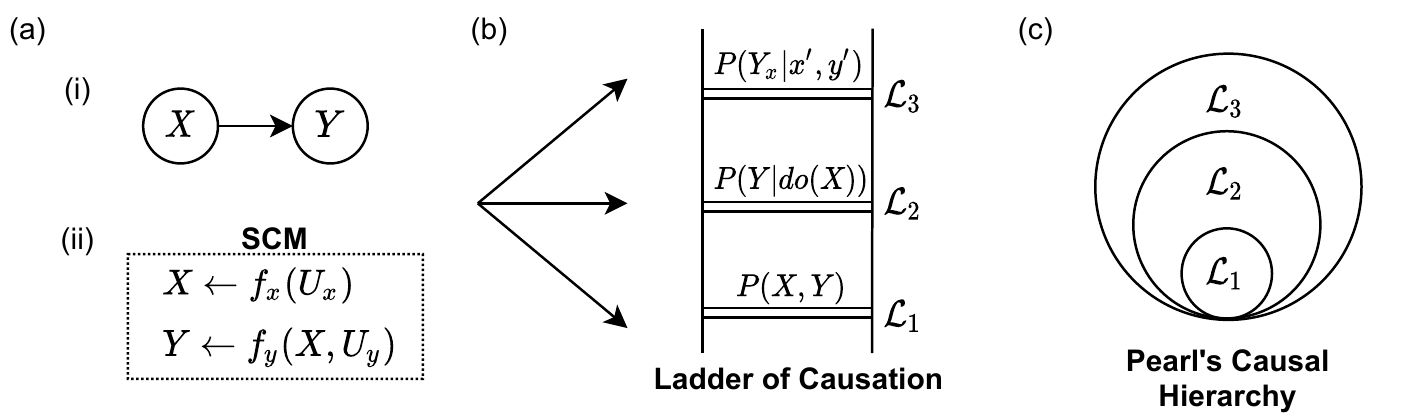}
    \caption{\textbf{(a)(i)} The \emph{causal Bayesian network}, represented by a DAG, depicting causal assumptions about variables and causal directions in a system. \textbf{(ii)} An example of a structural causal model (SCM) with two variables of interest corresponding to the DAG in (i). We say $X$ is \emph{exogenous} as it is completely determined by external noise factors, represented by $U_x$. $Y$ is \emph{endogenous} since it is determined by $X$ and some corresponding noise factor $U_y$.  \textbf{(b)} Shows the different types of observed phenomena induced by an SCM. The SCM induces three distinct types of causal quantities, each falling on a separate level of the `ladder of causation.' Associational (statistical) quantities fall on $\mathcal{L}_1$. Interventions (or experiments) correspond to forcing a variable $X$ to take on value $x$, thereby removing dependencies on parent variables. We represent this as $do(X=x)$ in Pearl's do-calculus, and these exist on $\mathcal{L}_2$.  Finally, counterfactual quantities are indicated by a distribution observed under one set of conditions $Y_x$, but where we now consider an alternative reality where $X=x^\prime$ such that $Y_x \mid x^\prime$. These exist on $\mathcal{L}_3$. \textbf{(c)} Venn diagram showing the containment relation described by Pearl's Causal Hierarchy \citep{ref:BCII-2020}, formalising the hierarchy the `ladder of causation' induces in (b).}
    \label{fig:SCM_Ladder}
\end{figure}

\textbf{Causality}. Causal inference considers how and when causal conclusions can be drawn from data. Recently, there has been great interest in benefits that derive from being explicit about causal assumptions in a modelling procedure. In general, complex systems of interacting variables can be described with a Structural Causal Model (SCM) \citep[][]{ref:Causality20090_Pearl,ref:ElementsOfCausalInference}. Such a model describes the causal mechanisms and assumptions present in an arbitrary system. The relationships between variables can be graphically presented in the form a \emph{causal Bayesian network} (CBN), wherein nodes represent variables and directed paths represent causal influence between variables (see Figure \ref{fig:SCM_Ladder}). The reader should not be be confused by related work in Structural Equation Modelling (SEM) which has roots in causal modelling \citep{ref:CausalOriginsOfSEMs}. \cite{ref:Pearl_GraphsCausalitySEMs} clarifies this confusion about causal assumptions in SEMs by explaining when such methods are valid for claiming causal outcomes. \cite{ref:ElementsOfCausalInference} formulates the SCM as follows:

\begin{definition}[Structural Causal Model (SCM)]
    A structural causal model $M = (S, P_{U_j})$ is a collection $S$ of $d$ structural assignments $X_j \leftarrow f_{j}(PA_j, U_{j})$, $j = 1,\dots,d$ where $PA_j$ are the parent nodes of $X_{j}$, and $P_{U}$ is a joint distribution over the product of jointly independent noise variables, $U_j$. An SCM implies a distribution $P$ with density $p$ over variables $X$ in the causal system.
\end{definition}

The key feature of an SCM, as opposed to an MDP, is that it allows for explicit consideration of counterfactual queries. Consider the "does smoking cause cancer?" debate. Perhaps it is possible that there is some hidden genetic factor that causes both cancer and the desire to smoke. Given that it isn't feasible to perform a randomised control trial, an agent must resort to other means of reasoning about this question. Pearl's \emph{ladder of causation} encapsulates the limitations of different systems of reasoning by separating them into three distinct classes \citep{ref:TheBookOfWhy, ref:SevenTools_Pearl}. Firstly, \emph{seeing} (associations) encapsulates statistical reasoning. The second rung corresponds with \emph{doing} (interventions), which contains state-of-the-art randomised control trials (RCTs) and RL methods \citep[][]{ref:RLisnotCausal, ref:BCII-2020}. More specifically, in RL an intervention corresponds to an experiment or \emph{action}, where an agent \emph{intervenes} on the natural state and \emph{causes} a response (e.g. next state and reward). One rung above knowledge accessible by intervention sits \emph{imagination} (counterfactuals). Such counterfactual queries are inherent to the RL problem where an agent could wonder whether it should have previously taken an alternative action. More simply, this formalism allows for explicit reasoning about each action an agent \emph{does}, \emph{can}, and \emph{could} have taken. The nuances of this `ladder' are formalised in Pearl's Causal Hierarchy \citep{ref:BCII-2020}, which establishes the containment relation between different types of interaction with the data-generating processes in a causal model. To be explicit, we provide definitions for interventions and counterfactuals in an SCM. 

\begin{definition}[Intervention]
    An intervention $I$ in an SCM $M$ entails changing some set of structural assignments in $M$ with a new set of structural assignments. Assume the replacement is on $X_k$ given by assignment $X_k = \tilde{f}(\tilde{PA}_k, \tilde{U}_k)$, where $\tilde{PA}_k$ are the parents in the underlying new DAG. This change in causal mechanism entails a new interventional distribution, $P_{do(I)}$ and corresponding density $p_{do(I)}$ over the variables of the SCM.
\end{definition}

\begin{definition}[Counterfactual]
    Consider SCM $M = (S, P_{U_{j}})$ over nodes $X$, with observations $x$. A counterfactual SCM is defined by replacing the distribution of noise variables with the noise associated with the realisation of variables $\boldsymbol{X} = \boldsymbol{x}$. This counterfactual noise term is written as a conditional probability $P_{U \mid X = x}$.
\end{definition}

With a model established and with knowledge of variables \emph{exogenous} and \emph{endogenous} to the system, methods of \emph{identification} can be used to establish the queries that can be answered within the current model. With this context in mind, we proceed by establishing some key benefits that causality brings to RL, building a context for future research in bridging causality with MARL.


\section{Causal Reinforcement Learning}
\label{sec:CRL}
Recent research has looked at tying together methods from causality with the high performance and powerful learning methods of RL, and is referred to as \emph{Causal Reinforcement Learning} (CRL) \citep[][]{ref:BareinboimTutorial,ref:CRL-InstrumentalVariablesApproach}. The structural invariances and performance guarantees gained from explicit causal reasoning, tied together with the learning ability of RL agents, allows for important problems to be considered with a learning-based approach. We now discuss some of the benefits that CRL has established.

\textbf{Transportability and Data Fusion}. A crucial problem in various data-driven machine learning domains is that of mismatched data. When data for learning is collected under mixed policies and environmental conditions, bias makes it difficult to establish rigorous, statistically significant results. Some examples include confounding shift \citep[][]{ref:ConfoundingShift}, various types of distribution shift \citep[][]{ref:OfflineRL-Levine,ref:MetaRL_RobustDistributionShift,ref:DistributionShift_OnlineRl}, and inductive biases \citep{ref:InductiveBiases}. By making assumptions about the data-generating system explicit, and reasoning about the causal relationships between variables of interest, it is possible to establish procedures for algorithmically combining datasets collected under different conditions and policies \citep{ref:ExternalValidity-Solved}. \cite{ref:BareinboimPearl-DataFusion} discuss criteria and techniques for combining datasets curated under different, causally related conditions modelled by an SCM.

A concept related to data fusion is that of \emph{transport}, which asks: when can we feasibly apply a policy learned under one set of conditions when operating under a different set of conditions? Causal inference can inform as to when and how results can be transported \citep[][]{ref:PB-2015_ExternalValidity}. A causal model of the underlying system allows for establishing bounds on performance, while also providing techniques for selecting optimal actions \citep[][]{ref:LB-2018, ref:LB-2020}.

\textbf{Generalised Off-Policy Learning}. As is thematic in machine learning in general, RL could benefit immensely from offline and off-policy methods that can learn from existing datasets \citep{ref:OfflineRL-Levine}. As we discussed in the context of data-fusion and transport, causal inference offers tools for correcting for biases by examining the effect of agent interventions under different policies \citep[][]{ref:InstrumentalVariables_Causality}. By reasoning causally, we can combine offline and online modes of learning to effectively improve upon policies - even when we only have observational data \citep{ref:NKYB-2020}. In contrast to conventional supervised and unsupervised methods, the \emph{learning} aspect of RL allows for extracting policies that perform beyond what model fitting of an offline dataset can yield. We can take this a step further by combining methods for offline RL with online environment interaction using a causal model. \cite{ref:ZB-2017} considers this generalised learning problem in the simplified setting of multi-armed bandits and show improved performance over conventional bandit methods. Multiple papers tackle healthcare problems, such as DTRs \cite[][]{ref:ZB-2019,ref:ZB-2020a,ref:NKYB-2020}, and medical diagnosis \citep[][]{ref:CausalML_MedicalDiagnosis_Nature}, where guarantees on performance are critical.

\textbf{Counterfactual Reasoning}. Beyond the previous stated benefits, the ability to reason about counterfactual queries is inherently useful. The ability to ask "what if?" is crucial for problems where experiments (taking action) are expensive, safety critical, or simply impossible. Improving the data-efficiency of learning algorithms will require making full use of available information for reasoning. Related to counterfactual reasoning, the idea of imagination within a world model has been studied in the context of RL \citep{ref:WorldModels}. Making such a model causal could provide interesting opportunities for \emph{counterfactual imagination}.

Counterfactual reasoning has motivated new paradigms for MDP-like sequential decision making problems. The MDP with Unobserved Confounders \citep{ref:BFP-2015, ref:ZB-2016}, models the setting where state, action, and reward variables are confounded by some latent external factors, potentially causing dependence which can harm learning performance. Further work has considered how consideration for \emph{intended} actions (i.e. the action that would have been taken) can reveal important information about confounders present in the system \citep{ref:ZB-2020c}. Especially relevant for RL, a counterfactual approach is also useful for directing exploration towards areas of the system that have causal unknowns and for optimising chosen actions (interventions) within the causal model  \citep[][]{ref:CausalMDP, ref:LB-2020}. These causal approaches are also being applied to imitation learning, where inference within a causal model allows for guarantees on behaviour \citep{ref:ZKB-2020}.

\textbf{Causal Learning}. The process of learning causal structure from data is a rich and active area of research \citep{ref:CausalDiscovery_GraphicalMethods, ref:JKSB-2020}. Most approaches rely on testing for conditional Independence, but considering other factors, such as time \citep{ref:CausalDiscovery_TimeSeries} and noise \citep{ref:CausalDiscovery_NonlinearAdditive}, has proven useful. Some common causal discovery methods include the PC and FCI algorithms \cite{ref:SpritesBook_PCAlgorithm}. The problem posed by causal discovery is related to problems posed in RL literature, where model-based approaches often propose learning a model before using it for planning \citep{ref:ModelBasedRL_Survey}. Further, RL methods have been applied for causal structure learning  \citep{ref:CausalDiscovery_RL}. The authors believe there is great opportunity for merging model-based RL and causal discovery methods.

\section{The Multi-Agent Case}
Multi-Agent Systems (MAS) is an area of research that studies the interaction of intelligent agents \citep{red:IntroToMAS}. As in RL, multi-agent learning is studied under various models, often extending the RL paradigm to multiple agents \citep[][]{ref:TheoryOfGames_ExtensiveForm_Def,ref:MarkovGames,ref:MARLSurvey2003,ref:MARLSurvey2021}. A key benefit of the multi-agent approach is the decentralisation of the learning task, which may more naturally map to many potential RL applications. While there has been some interest in merging ideas from causality with current single-agent methods, to the best of our knowledge no research explicitly looks at bridging (graphical) causal methods with MARL. Prior work has looked at exploiting counterfactual knowledge for the multi-agent cooperative setting. For example, \cite{ref:COMA} improve performance on tasks where credit assignment is challenging by using counterfactual information. \cite{ref:CausalCommunication} also consider the credit assignment problem, improving upon previous results by communicating using counterfactuals. \cite{ref:SocialInfluence_Communication} propose using causality for intrinsic motivation via social influence. There has also been broader interest in counterfactuals outside of RL. For example, MAS literature considers counterfactuals in terms of \emph{wonderful life utility} \citep{ref:WolpertTumer_WonderfulLifeUtility}. Related work also looked at \emph{difference rewards} as a way to reduce noise and allow agents to learn the `consequences of their actions' \citep{ref:DifferenceRewards}.

Causal models are especially relevant in the context of high performance, deep-learning methods which have only recently used for MARL problems \citep[][]{ref:QMIX, ref:RIAL_DIAL, ref:Grandmaster_StarCraft, ref:MARL_NetworkedAgents}. Such multi-agent paradigms immediately afford researchers opportunity to exploit the nature of agent cooperation and interaction, including imitation and communication. Additionally, multiple agents comes with inherent scalablity benefits due to decentralisation of learning and/or execution \citep{ref:ScalableMARL,ref:RobustScalableMARL,ref:MAVA}. It is our belief that causal methods are especially well suited for dealing with the additional challenges and complexity in multi-agent problems. These complexities are reflected in the various models one finds in the MARL and game theoretic literature \citep[][]{ref:MarkovGames, ref:DynamicProgramming-POSG,ref:Dec-POMDP,ref:TheoryOfGames_ExtensiveForm_Def, ref:Merging_POSG_EFG,ref:MultiObjective_MultiAgent}. With that said, most modern MARL papers formulate the MARL problem as some multi-agent extension of the MDP. One such extension is the Stochastic (or Markov) Game, where the system evolves in discrete steps according to the combination of state-actions selected by agents in the environment. We can extend stochastic games to the partially observable case where each agent now only has access to some function of the true state (an observation) \citep{ref:DynamicProgramming-POSG}. In the cooperative joint reward and shared history setting, the Partially Observable Stochastic Game (POSG) simplifies to the \emph{decentralised} POMDP (Dec-POMDP) \citep{ref:Dec-POMDP}. We provide Definition \ref{def:Dec-POMDP} below to highlight the similarity to the definition of the MDP (Definition \ref{def:MDP}). For the sake of clarity of exposition we focus on this simplified setting for the remainder of this paper. With these models in mind, we proceed by shifting our focus to a `causality first' perspective for multi-agent problems and discuss how this may advance research in the area.

\begin{definition}[Dec-POMDP]
\label{def:Dec-POMDP}
    A Dec-POMDP is a multi-agent extension of a POMDP. It is a 7-tuple $\left\langle S,\left\{A_{i}\right\}, T, R,\left\{\Omega_{i}\right\}, O, \gamma\right\rangle,$ where $S$ are states, $\left\{A_{i}\right\}$ is the joint action set, $T = P(s^\prime \mid s, a)$ is the set of conditional transition probabilities between states, $R$ is the reward function, $\left\{\Omega_{i}\right\}$ is the joint observation set, $O(s^\prime,a,o)=P(o \mid s^\prime,a)$ gives the conditional observation distribution, and $\gamma \in [0,1]$ is the discount factor.
\end{definition}

The multi-agent causal problem has also been considered from a Bayesian learning perspective. The Bayesian Game \citep{ref:BayesianGame} is a natural formulation for situations where agents have limited information about the actions that other agents take, as well as the reasons they take those actions - their policies. \cite{ref:BayesianCausalGames} extend this idea by including a causal model about the environment. Agents must then maintain a belief about the causal nature of the system. The Bayesian approach leads to a rational decision making criterion, where agents hold probabilistic beliefs about possible causal models of the environment. The game theoretic approach leads to the concept of a \emph{causal Nash equilibrium}, where cooperative agents jointly optimise their actions. The authors feel this is an unexplored and ripe opportunity for application to the MARL problem.

The difficulty of scaling Dec-POMDPs has prompted methods that exploit local dependencies between different agents and their environment. For example, factored Dec-POMDPs \citep{ref:FactoredDecPOMDP} extend Dec-POMDPs by including \emph{factors}, defined as $\mathcal{X} = \{\mathcal{X}_1,\dots,\mathcal{X}_{|\mathcal{X}|}\}$ that span the state space of the Dec-POMDP. A state is then an assignment of these factors, $s=\{x_1,\dots,x_{|\mathcal{X}|}\}$. Conditional independence relations between variables in the factored model can be exploited to reduce the dimensionality of the learning problem. \cite{ref:FactoredDecPOMDP} originally suggested using a dynamic Bayesian Network (DBN) to model causal influence in the transition and observation models. The ability for the factored approach to build up an interaction model for the system means it is causally aware at the level of interventions $\mathcal{L}_2$. For the full realisation of causal RL benefits, we really want agent models to be capable of counterfactual reasoning. 

\emph{Multi-Agent Causal Models} (MACM) were proposed in the context of extending causal models to the decentralised multi-agent case, where agents share an environment and have access to private and/or public variables of interest \citep{ref:MultiAgentCausalModels}. \cite{ref:MACM_Learning} use MACMs for distributed structure learning, showing that it can be effective in multi-agent causal reasoning problems. The distinguishing factor in MACMs, as opposed to MARL models, is a lack of explicit formalisation of an RL-like learning process. This difference parallels the difference in SCMs and single-agent RL models (see Section \ref{sec:CRL}). This motivates considering reframing common MARL models in terms of MACMs, perhaps allowing for methods that can reason about counterfactual trajectories.

\begin{definition}[Multi-Agent Causal Models]
    A multi-agent causal model (MACM) is a set of $n$ agents, each having access to a semi-Markovian model $M_i$ defined as $M_i = \langle V_{M_i}), G_{M_i}), P(V_{M_i}), K_{M_i} \rangle, \quad i \in \{1,
    \dots, n\}.$ Here $V_{M_i}$ represents the model variables agent $i$ has access to, while $G_{M_i}$ is the causal DAG over $V_{M_i}$. $P(V_{M_i})$ is the joint distribution over $V_{M_i}$. Finally, $K_{M_i}$ indicates which variables are shared between agent $i$ and other agents $j$.
\end{definition}

\section{Towards Causal Multi-Agent RL}
\label{sec:CausalMARL}
\begin{figure}[!b]
    \centering
    \includegraphics[width=\linewidth]{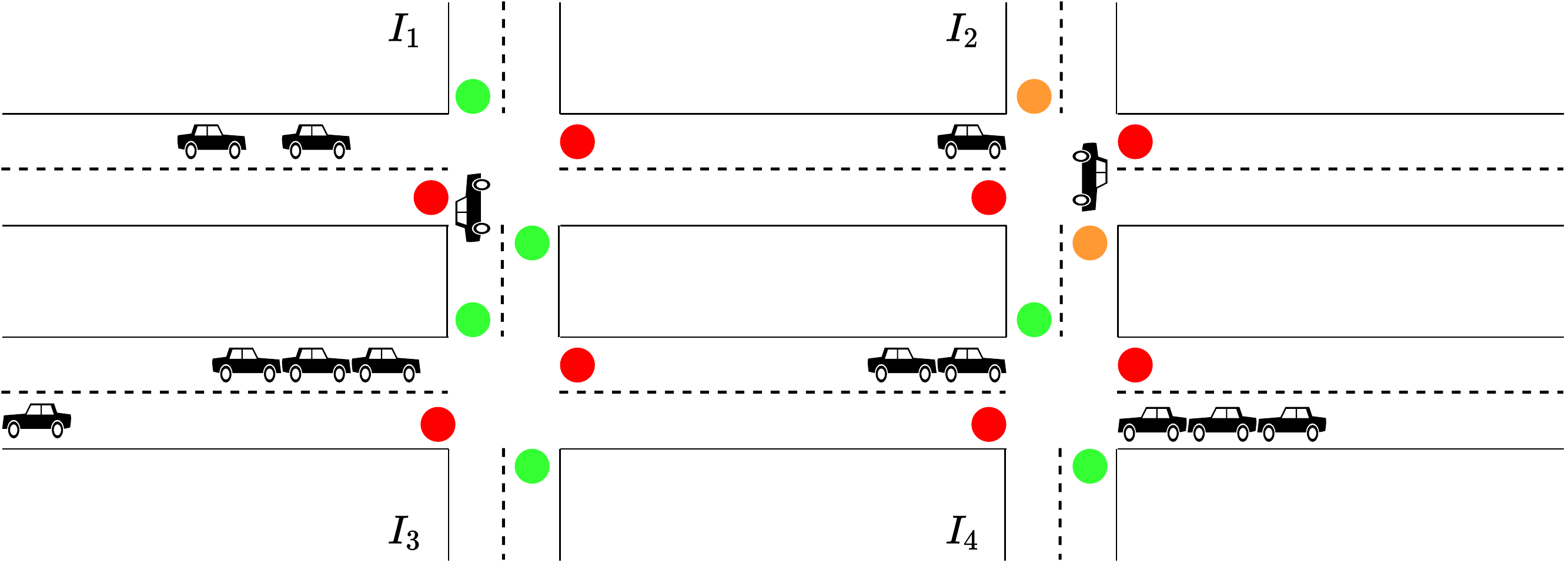}
    \caption{Figure showing system of four intersections, $\{I_1,\dots,I_4\}$, each treated as an agent in the \emph{traffic light control example} of Section \ref{sec:CausalMARL}. Agents try to minimise global traffic by controlling traffic lights given only measurement of traffic on adjacent roads. Best viewed in colour.}
    \label{fig:intersections}
\end{figure}

In this section we examine a way to frame a Dec-POMDP as a multi-agent causal model. We consider the simplified two-agent setting where both agents have limited access to knowledge of the true environment state in the form of observations, but we note that the MACM can be used in a similar way to model more general cooperative scenarios. We discuss one such example where observations are only shared between subsets of agents (see Figure \ref{fig:intersections}).

Returning to the two-agent setting, by modelling state-observation-action trajectories as structural assignments in a causal model we can form two independent SCMs. This means we can be explicit about how the models of interacting agents are related in terms of the components that are shared amongst them. We note that agents share access to the environment and experience the same global underlying state, while also sharing access to the history of state-action trajectories of all agents. In this way, we have a Dec-POMDP modelled as a MACM. We note that although this does not differ substantially from the SCM view of the Dec-POMDP, an MACM formulation handles the sharing of agent variables explicitly whereas an SCM does not. This is useful, for example, in modelling more general scenarios where agent dynamics differ from the Dec-POMDP formulation. Consider the more general scenario where agents must cooperate to achieve optimal outcomes without a shared reward. In this sense, MACMs provide a `causal wrapper' for general classes of multi-agent models. With this said, we maintain focus on the Dec-POMDP example as it is a foundational model in MARL and is useful for exposition.

\begin{figure}[!t]
    \centering
    \includegraphics[width=\linewidth]{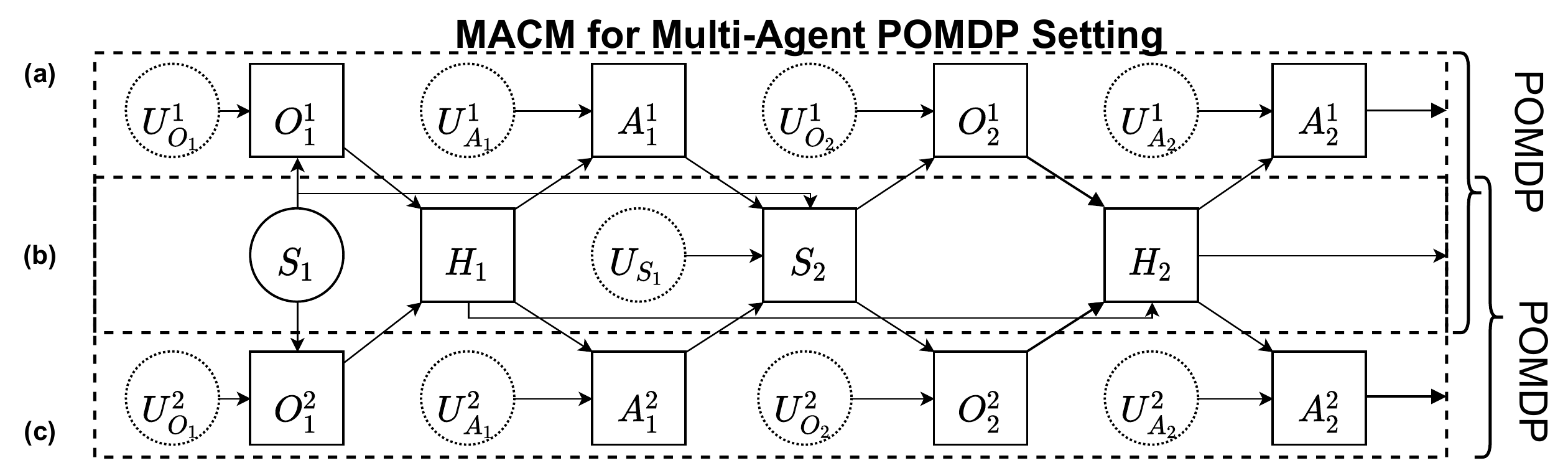}
    \caption{Diagram representing a two-agent sequential decision making system viewed through the lens of an MACM. Squares are used to represent endogenous variables and their associated structural causal assignments. Circles represent exogenous variables, with noise variables highlighted by dotted lines. (a) and (c) can be viewed as individual POMDPs of the agents represented as a Bayesian causal network with noise variables explicitly represented. $S$ represents the global state of the system, $O$ represents observations of the individual agents, $H$ depicts the shared histories (potentially previous observations, actions, and/or rewards depending on problem setting), and $A$ represents actions selected by the individual agents. Agent variables are indicated by superscripts. Noise variables are represented by $U$. (b) highlights overlap in the POMDPs. The system is an MACM, but we note MACMs can be similarly used to model more general MARL models.}
    \label{fig:CD-POMDP}
\end{figure}

To formalise modelling a Dec-POMDP as an MACM, we consider all conditional distributions in the graphical model as being deterministic functions with associated independent noise $U$ (see Figure \ref{fig:CD-POMDP}). This includes the transition functions such that a state is causally determined by the previous state, actions and relevant noise terms, $S_{t+1} = f_{st}(S_t, A^1_t, A^2_t, U_{S_t})$. Here $S_t$ represents the environment state at time-step $t$, $A^1_t$ represents actions taken by agent 1 at time-step $t$, and $U_{S_t}$ is the noise associated with the determination of the underlying state. 

\textbf{Traffic Light Control Example.} Consider the four interlinked intersections shown in Figure \ref{fig:intersections}. The goal here is to minimise the total traffic experienced by drivers without full state information. For scaling reasons, we would like to have a decentralised system where each intersection controls the traffic lights given only information about the traffic level on adjacent roads. Modelling this scenario as a Dec-POMDP is possible but is perhaps more efficiently represented in a factored manner. Similarly to the factored Dec-POMDP approach, the decentralised nature of MACMs means we assign each agent an SCM representing its causal model. Crucially, we can explicitly model the shared variables between the neighbours (intersections denoted by $I_i$) in an efficient manner. For example, $I_1$ shares traffic observations with both the $I_2$ and $I_3$. However, $I_2$ and $I_3$ share no observations. The main benefits of this causal modelling approach is that it opens the door for methods that rely on the causal assumptions applied in SCM (and by extension MACM) modelling. We now discuss some specific potential benefits we see emerging in future research in causal MARL.

\textbf{Tackling Non-stationarity}. Individual agents face a moving target problem when other agents simultaneously adapt their policies in response to observed outcomes. This \emph{non-stationarity} arises from limited information about other-agent policies, which is fundamental to the multi-agent, decentralised paradigms \citep[][]{ref:MARL_Non-Stationarity_Survey}. Common approaches to tackling this include centralising the training procedure \citep{ref:MARL_CentralisedTraining_Survey}, accounting for other agents \citep{ref:ModellingOthers_MARL}, and applying communication protocols (see section on \emph{Knowledge Sharing} below). Fundamentally, non-stationarity is a source of uncertainty for individual agents. Causal inference offers tools for modelling such uncertain scenarios, especially where agents have structural knowledge of \emph{how} uncertainties are arising \citep{ref:LB-2020,ref:Ignorance_Confounding_Estimation}. Related work has been considering this problem in the Dec-POMDP setting as `multi-agent beliefs' \cite[see p.59][]{ref:DecPOMDPs_Textbook}. For example, an agent may know that there is another agent in its vicinity, and it can therefore attribute (some of) the noise in its observations to the unknown behaviour policy of those agents. 

\textbf{Knowledge Sharing}. An effective way to increase cooperation in multi-agent systems is by using communication methods \citep{ref:RIAL_DIAL, ref:SocialInfluence_Communication, ref:TarMAC_Communication, ref:IntentionSharing}. Such communication methods are designed to boost learning performance by exploiting existing knowledge encoded in other-agent networks. These communication methods are often less natural than real-world emergent behaviour displayed by human communication, and there is room for research for emergent human intelligible and interpretable communication. One can imagine that communicating causal relationships could greatly boost learning efficiency in decentralised tasks, similar to how \cite{ref:CausalCommunication} improve credit assignment using counterfactual knowledge. This is potentially a fruitful direction for multi-agent, causal research.

Another promising method of knowledge sharing is by applying transfer learning and imitation learning. In these settings, a more knowledgeable agent shares or teaches acquired knowledge to another agent \citep{ref:TransferLearning_Survey}. An exciting CRL research direction is that of causal imitation learning where a causal model of the environment can be used to determine when imitation is feasible, and what performance we can expect. Reasoning causally has shown to aid in determining what variables are important for the imitation procedure \citep[][]{ref:ZKB-2020}. Tying these ideas together with (causal) transfer learning and data-fusion opens room for exciting research avenues. For example, teacher-student learning \citep{ref:TeacherStudentFramework, ref:LearningToTeach_CoopMARL, ref:TeachingOnBudget}, emergent behaviour \citep{ref:ROMA_EmergentMARL}, and general cooperative learning could benefit from such ideas. 

\textbf{Decentralised Reasoning}. Embedding a causal model in the multi-agent framework opens opportunity for decentralised reasoning in safety critical applications such as healthcare and robotic applications. For example, consider how healthcare practitioners can pool resources \citep{ref:FederatedHealthcare} and optimise their healthcare policies aided by a safe and interactive MARL systems \citep{ref:InteractiveML}. Bridged with other benefits of explicit causal methods already discussed, this presents a promising front for future research. Other topics of interest include multi-agent modelling of market dynamics \citep[][]{ref:MARL_StockMarket}, multi-agent bidding in advertising \citep[][]{ref:AdBiddingMARL}, and traffic system control \citep{ref:MARL_TrafficControl}.

\textbf{Generalised Off-Policy MARL}. The primary challenge with offline RL is that we are limited to observational data. Viewed from another angle, this problem comes down to learning structure from data collected under mixed policies and is subject to distribution shift (see Section \ref{sec:CRL}). Explicitly modelling the assumptions and uncertainties can aid in bounding the uncertainties in the learning process \citep{ref:PropensityScore, ref:BoundingCausalEffects}. Extracting maximal information from data is of primary interest in the causal learning literature, and RL researchers interested in this overlap will only benefit as new results emerge. The distribution shift problem is especially exacerbated in the multi-agent problem, where multiple agents with hidden decision policies are influencing the response of the environment \citep{ref:OfflineMARL-ImplicitConstraint, ref:OfflineRL-Levine, ref:MARL_Non-Stationarity_Survey}. Beyond the fully offline case, many applications could benefit by improving policies with existing offline datasets. This differs from conventional off-policy learning because we lack knowledge of how data was collected. MARL extends the difficulty of this generalised setting by involving more decision agents. We believe this is an interesting research direction.

\textbf{Model Learning}. As in model-based (MA)RL, it may be of interest to first learn a causal model and then use it for identification and planning. Recent causality literature usually frames this as \emph{causal discovery}. These methods have been applied in practice, yielding some impressive results \citep{ref:CausalDiscovery_NonlinearAdditive, ref:CausalDiscovery_RL, ref:CausalDiscovery_GraphicalMethods}. Structure learning of multi-agent models has been considered in \cite{ref:MACM_Learning}. This presents interesting opportunities for research in MARL where, for example, MARL could be used as a heuristic for learning direction of causality by rewarding an agent for correctly choosing actions so as to increase knowledge useful for structure learning.

\section{Discussion}
In this paper we have introduced selected ideas from causal inference and causal RL with the intention of promoting interest in applying similar methods for MARL. We have discussed a broad range of topics where causality appears to provide interesting avenues for research, including how explicit modelling of causal assumptions can bridge mixed modes of data collection and tackle problems in offline MARL. The authors would like to point out that causal modelling approaches are not limited to graphical model literature, though we have mostly focused on these here as graphical methods are well suited to current approaches to computational sciences in addition to being easy to understand. 

There exists extensive literature on \emph{potential outcomes} and \emph{propensity scores} \citep[][]{ref:RubinCausalModel, ref:PropensityScore}. \emph{Instrumental variable} \citep[][]{ref:InstrumentalVariables_Causality} approaches are very common in practice, recently extended as a basis for CRL \citep{ref:CRL-InstrumentalVariablesApproach}. To maintain perspective on the current state of causal inference literature, we point to a critiques of SCMs \cite[e.g.][]{ref:SCM_Critique}. We also make note of relationships between differential equations and causal models, which are investigated in various settings \citep{ref:ODEs_SCMs, ref:RandomDEs_SCMs, ref:CausalityForMachineLearning}.

\subsection*{Acknowledgement} St John Grimbly acknowledges support from the University of Cape Town and ETDP SETA.

\newpage

\bibliographystyle{IEEEtranN}
\bibliography{references}



\end{document}